\documentclass[10pt,twocolumn,letterpaper]{article}

\usepackage{cvpr}              

\usepackage{bbding}
\usepackage{colortbl}
\usepackage{float} 
\definecolor{cvprblue}{rgb}{0.21,0.49,0.74}
\usepackage[pagebackref,breaklinks,colorlinks,allcolors=cvprblue]{hyperref}

\usepackage{algorithmic}
\usepackage[ruled,linesnumbered]{algorithm2e}

\def\paperID{2484} 
\def\confName{CVPR}
\def\confYear{2025}

\title{FlowRAM: Grounding Flow Matching Policy with Region-Aware Mamba Framework for Robotic Manipulation}

\author{
Sen Wang$^{1}$ ~Le Wang$^{1, \text{\Envelope}}$ ~Sanping Zhou$^1$ ~Jingyi Tian$^1$ ~Jiayi Li$^1$ ~Haowen Sun$^1$ ~Wei Tang$^2$\\
$^{1}$National Key Laboratory of Human-Machine Hybrid Augmented Intelligence, \\ 
National Engineering Research Center for Visual Information and Applications, \\ 
Institute of Artificial Intelligence and Robotics, Xi'an Jiaotong University \\
$^{2}$University of Illinois at Chicago
}

\begin{document}
\maketitle
\begin{abstract}

\noindent Robotic manipulation in high-precision tasks is essential for numerous industrial and real-world applications where accuracy and speed are required. Yet current diffusion-based policy learning methods generally suffer from low computational efficiency due to the iterative denoising process during inference. Moreover, these methods do not fully explore the potential of generative models for enhancing information exploration in 3D environments. In response, we propose FlowRAM, a novel framework that leverages generative models to achieve region-aware perception, enabling efficient multimodal information processing. Specifically, we devise a Dynamic Radius Schedule, which allows adaptive perception, facilitating transitions from global scene comprehension to fine-grained geometric details. Furthermore, we integrate state space models to integrate multimodal information, while preserving linear computational complexity. In addition, we employ conditional flow matching to learn action poses by regressing deterministic vector fields, simplifying the learning process while maintaining performance. We verify the effectiveness of the FlowRAM in the RLBench, an established manipulation benchmark, and achieve state-of-the-art performance. The results demonstrate that FlowRAM achieves a remarkable improvement, particularly in high-precision tasks, where it outperforms previous methods by 12.0\% in average success rate. Additionally, FlowRAM is able to generate physically plausible actions for a variety of real-world tasks in less than 4 time steps, significantly increasing inference speed.

\renewcommand{\thefootnote}{\text{\Envelope}}
\footnotetext{Corresponding author.}

\vspace{-2ex}
\end{abstract}    
\section{Introduction}
\label{sec:intro}

\begin{figure}[t]
    \centering
    \includegraphics[width=\linewidth]{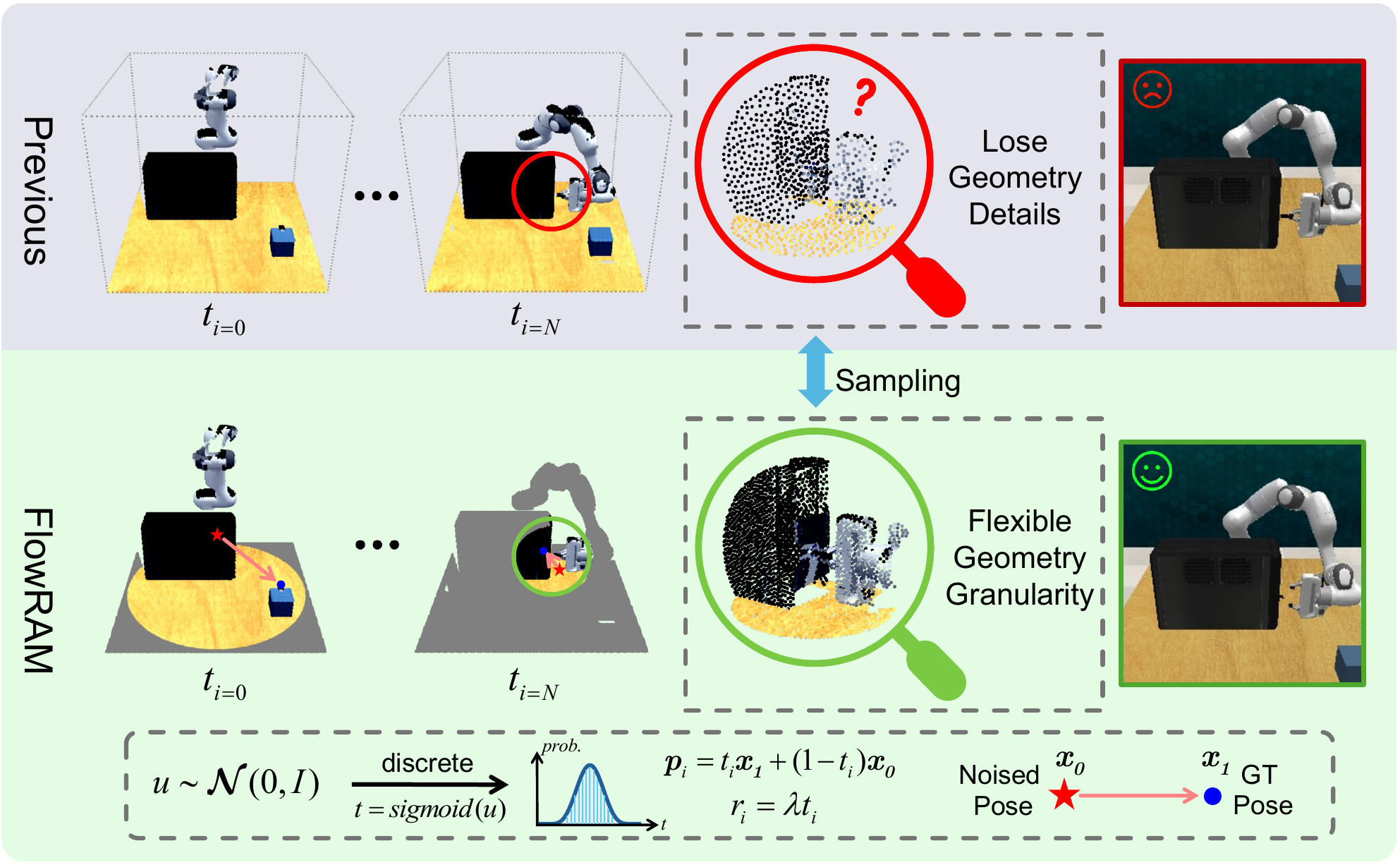}
    \caption{\textbf{The diagram of the proposed FlowRAM.} Consider the \texttt{insert USB in computer} task. Previous arts have attempted to capture information about the geometric details of USB and ports, but fail severely due to lacking task-relevant perception. In contrast, FlowRAM equipped with a dynamic radius schedule successfully preserves these geometric details and ensures accurate alignment of the USB and port. The \(\boldsymbol{p_i}, r_i\) in the figure represent the center and radius of the region, respectively.}    
    \vspace{-2ex}

    \label{fig:motivation}
\end{figure}


Designing agents for high-precision manipulation tasks has attracted increasing attention in the field of robotics and embodied AI~\cite{eisnerdeep,goyal2024rvt2,lloyd2024pose,zhao2023learning,burns2024genchip}. To succeed in such tasks, agents must not only capture comprehensive semantic information about the environment but also have a fine-grained geometric understanding of task-relevant regions. This level of detail is essential, as even small errors in perception or execution can result in significant failures in high-precision tasks.

To address these challenges, previous arts have advanced in both perception and policy learning. Regarding perception, 3D-based methods have been widely adopted for their ability to accurately infer precise 3D positions and spatial relations~\cite{hong20233d,chen2020scanrefer,Ze2024DP3,gervet2023act3d}. In terms of policy learning, prevailing methods can be divided into two categories. The first category comprises deterministic policies, which directly predict actions by outputting conditional probability distributions. These distributions represent the likelihood of different actions based on the current state or observations~\cite{james2022coarse,shridhar2023perceiver,chen2023polarnet,lu2024manigaussian,ze2023gnfactor,goyal2023rvt,urain2024deep,xia2022learning}. However, deterministic policies struggle with handling high-dimension distributions and scaling to complex action spaces. In contrast, policies based on generative models are particularly adept at navigating the intricate of high-dimensional continuous action spaces~\cite{chi2023diffusion,chernova2007confidence,mandlekar2020iris}. 
Notably, diffusion-based policies generate complex actions by iteratively removing noisy component through a denoising process to transform pure noise into realistic actions. This approach has witnessed a significant advancement in modeling continuous multimodal action distributions, allowing for diverse and effective action generation~\cite{chi2023diffusion,Ze2024DP3,ma2024hierarchical,xian2023chaineddiffuser}.

Despite notable progress, existing diffusion-based methods still face two fundamental challenges. First, they lack specialized perception of task-relevant regions, limiting their performance in high-precision tasks. Second, conventional diffusion models are inherently slow in generation due to their iterative denoising process, which restricts the applicability of diffusion policies for fast robot control.

To address these challenges, we introduce FlowRAM, a novel framework that exploits the process of conditional flow matching to introduce a region-aware strategy in a simple yet effective way, which improves both the precision and efficiency of robot manipulation.

\textbf{Precise 3D Region-Aware Perception with Flexible Geometry Granularity.} To achieve high precision, FlowRAM incorporates a 3D region-aware strategy that dynamically narrows its focus on task-relevant regions as the flow matching generation process unfolds. This is achieved through a dynamic radius schedule, where the perception range decreases progressively across discretized time steps, allowing FlowRAM to capture fine-grained geometric details essential for high-precision tasks, as showcased in~\cref{fig:motivation}. Unlike previous approaches overlook the local information, FlowRAM achieves flexible geometric granularity by sampling within these regions, enabling precise adaptation to the spatial intricate of each task-relevant region. A point cloud encoder~\cite{liang2024pointmamba} processes the sampled points, producing detailed features that enhance task accuracy. Furthermore, FlowRAM employs a standard Mamba model~\cite{mamba} for efficient multimodal fusion, which offers lower computational overhead and higher efficiency through its linear complexity. Its capability to flexibly capture and process detailed 3D features enables FlowRAM to complete tasks requiring precise manipulation, such as \texttt{insert USB in computer} or \texttt{screwing a nail}.

\textbf{Efficient Action Generation with Conditional Flow Matching.} To deal with the slow diffusion process, FlowRAM leverages conditional flow matching~\cite{liuflow,liu2022rectified,esser2024scaling,lipmanflow,chen2024flow} to generate a 6-DoF action pose directly by learning a parameterized velocity field that guides a random initial point toward the target. This approach simplifies the iterative denoising process required by conventional diffusion models, resulting in an efficient generation process which requires fewer time steps during inference. Starting from a noise distribution, FlowRAM learns the multimodal distribution of keyframe actions by fitting a vector field along probability flow paths, ensuring both stability and efficiency in generating actions.

In summary, the main contributions of this study can be summarized as follows:
\begin{itemize}
    \item We propose a novel 3D region-aware strategy that progressively focuses on task-relevant regions through a dynamic radius schedule, enabling precise feature capture essential for high-precision tasks.
    \item We incorporate conditional flow matching to achieve fast and precise keyframe pose generation. Additionally, we leverage state space models~\cite{mamba} as a multimodal fusion module, achieving superior performance with minimal computational cost.
    \item Extensive evaluations, including multi-task experiments across 10 tasks from RLBench~\cite{james2019rlbench} and 7 high-precision tasks~\cite{guhur2023instruction}, demonstrate that our method achieves new state-of-the-art performance in both settings. We also validated the effectiveness of FlowRAM through real-world deployments.

\end{itemize}

\section{Related Work}
\label{sec:related}


\subsection{Scene Perception for Robotic Manipulation} 
Early works~\cite{jang2022bc,brohan2022rt,shridhar2022cliport,driess2023palm,guhur2023instruction,reed2022generalist} primarily used 2D images for policy learning, but struggled to capture depth and geometry information, limiting their ability to handle complex tasks. To overcome these challenges, recent advancements in the field have incorporated 3D scene perception~\cite{chen2023polarnet,james2022coarse,shridhar2023perceiver,yuan2023m2t2,ze2023gnfactor,lu2024manigaussian,goyal2023rvt}, enhancing spatial reasoning by providing richer geometric insights into the environment, \eg, C2F-ARM~\cite{james2022coarse} and PerAct~\cite{shridhar2023perceiver} voxelized the workspace, detecting index of the voxel that contains the next keyframe action for the end-effector. ManiGaussian~\cite{lu2024manigaussian} models spatiotemporal scene dynamics to predict future states for precise robotic manipulation. RVT~\cite{goyal2023rvt} proposes a multi-view transformer architecture, leveraging re-rendered novel virtual views to enhance action prediction. However, these methods either rely on high-resolution voxel grids or fail to adequately capture critical regions for precise manipulation. Recent research~\cite{james2022coarse,gervet2023act3d,goyal2024rvt2} highlights that effective scene perception, particularly the identification of task-relevant regions, is crucial for improving performance in robotic manipulation tasks. Act3D~\cite{gervet2023act3d} employs a coarse-to-fine strategy for 3D point sampling, enabling variable spatial resolution. Nevertheless, deterministic policy learning makes it hard to generate various actions. RVT-2~\cite{goyal2024rvt2} adopts a two-stage network that first predicts a region of interest, then refines the prediction with virtual rendering. However, the rendered views are susceptible to occlusions, and also limited in their ability to sample consistent actions from predicted heatmaps.

Compared to the previous arts, we leverage the flow matching process to dynamically adjust the 3D perception, effectively focusing on task-relevant regions of the workspace. Additionally, FlowRAM follows the keyframe-based manipulation paradigm~\cite{goyal2023rvt,goyal2024rvt2}.

\subsection{Flow Matching for Policy Learning}
Diffusion models approximate data distributions via an iterative denoising process and have achieved significant success in image and video generation~\cite{song2020denoising,li2024diff,esser2024scaling}. In robotics, these models are employed to learn action score functions conditioned on observations, language, and robot states, enabling the generation of multimodal trajectories or actions~\cite{ma2024hierarchical,ke20243d,Ze2024DP3,oba2024read}. However, existing diffusion-based models rely on an iterative denoising process, resulting in extremely low efficiency. Where flow matching~\cite{lipmanflow,dao2023flow,davtyan2023efficient}, as a novel generative model training method, establishes a direct connection between noisy and target data through deterministic straight-line paths in vector fields, and has shown superior performance over diffusion models in tasks such as trajectory generation~\cite{zhang2024affordance,chisarilearning,funk2024actionflow}, and decision-making~\cite{braun2024riemannian}. In this work, we exploit Conditional Flow Matching (CFM) to parameterize a velocity field and then numerically integrate along the flow path to accurately predict the next keyframe pose. Furthermore, extensive experiments show that CFM outperforms established diffusion objectives in the field of robotic manipulation.

\subsection{State Space Models in Robotics}
State Space Models (SSMs) have emerged as a powerful alternative to traditional CNNs and Transformers~\cite{mamba,gu2022efficiently,nguyen2022s4nd}, offering notable advantages in efficiency and scalability. Among these, the Mamba model~\cite{mamba} stands out with its selective SSM mechanism and a hardware-aware algorithm that enables linear-time inference and efficient training. These capabilities have attracted widespread attention, establishing Mamba as a versatile framework across various fields~\cite{guo2024mambair,dong2024hambasingleview3dhand,lin2024mtmamba}. Building on these strengths, research has explored the applications of Mamba in visual representation learning~\cite{liu2024vmamba,vim}, point cloud processing~\cite{liang2024pointmamba,zhang2024point}, and robotic manipulation~\cite{liu2024robomamba,cao2024mamba,ma2024madiff}. While previous works have employed Mamba as either a vision processor or a policy module in robotic manipulation tasks, our work develops an integrated Mamba framework. This unified approach combines both perception and policy, optimizing system efficiency and enhancing coordination for more robust robotic control.

\begin{figure*}[t]
    \centering
    \includegraphics[width=1\linewidth]{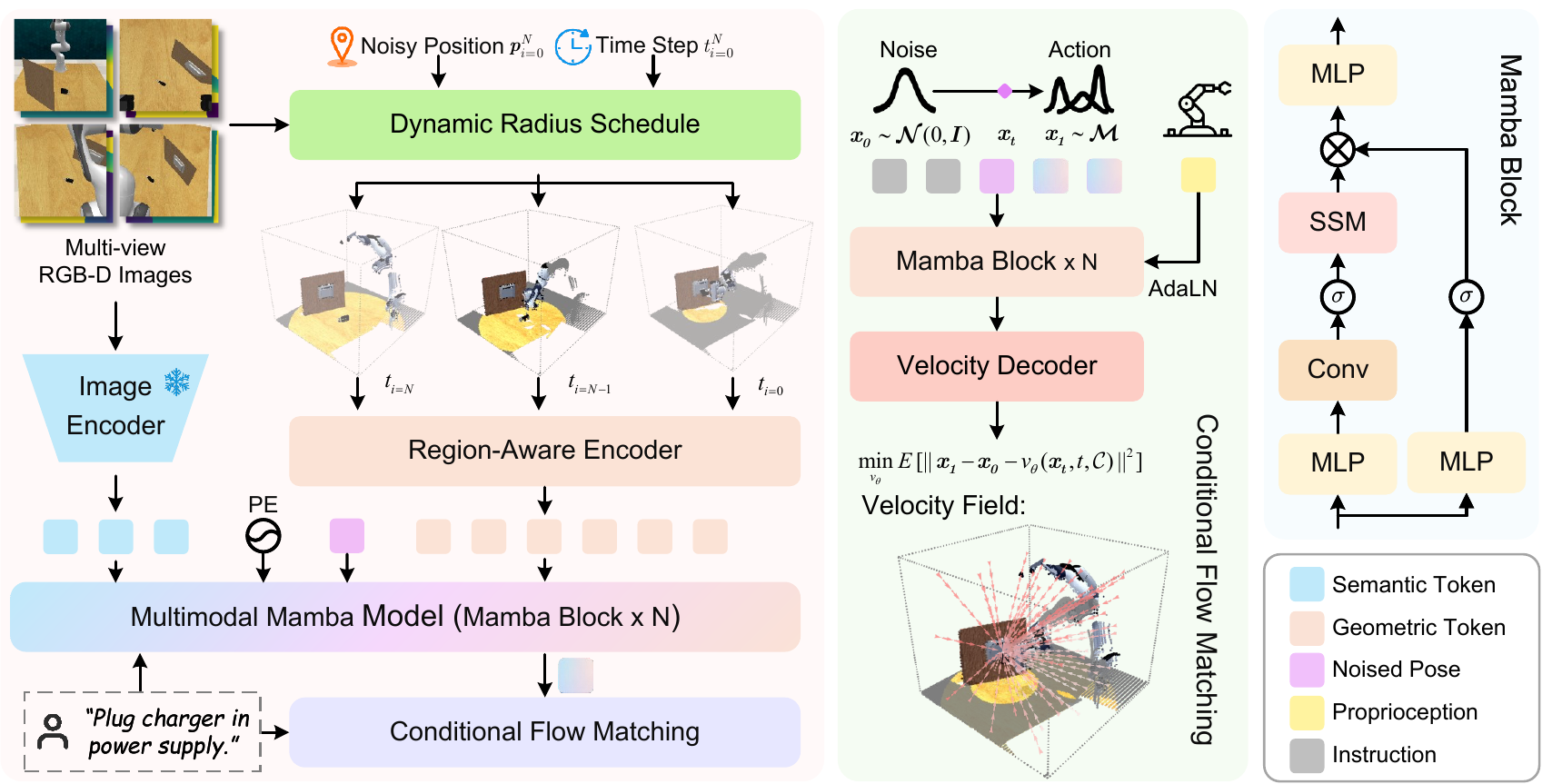}
    \caption{\textbf{The overall framework of \textit{FlowRAM}.} FlowRAM is a Mamba-based framework, where multimodal Mamba processes multi-view RGB images, geometric information, instructions, and robot proprioception. In flow matching, noise-perturbed actions are transported to target actions guided by observations. During training, a dynamic radius schedule adjusts the perception radius over random discrete time steps, enabling the model to capture regions of varying sizes and adapt to diverse manipulation tasks by extracting comprehensive semantic and geometric features. }
    \label{fig:pipeline}
    \vspace{-2ex}
\end{figure*}

\section{Method}
\label{sec:Methodology}

In this section, we present FlowRAM, a novel Mamba-based framework that fully exploits the flow matching generation process for efficient robot learning. We commence with a brief background on conditional flow matching~\cite{liuflow,liu2022rectified} and state space models~\cite{mamba}. We then delve into the key designs of FlowRAM, including the Dynamic Radius Schedule (DRS) for region-aware perception, followed by key architectural components and implementation details. The overall framework of FlowRAM is illustrated in~\cref{fig:pipeline}.

\subsection{Preliminaries}

\textbf{Conditional Flow Matching.} 
Conditional Flow Matching (CFM) is a flow-based method that aims to transport data between two probability distributions \( \pi_0 \) and \( \pi_1 \) using an Ordinary Differential Equation (ODE) framework. CFM learns a discrete-time ODE model that transfers samples \( \boldsymbol{x}_0\) from initial distribution \( \pi_0 \) to \( \boldsymbol{x}_1 \) from target distribution \( \pi_1 \). The learned ODE governing the data transfer is defined as:
\begin{equation}
    d\boldsymbol{x}_t = \boldsymbol{v}_\theta(\boldsymbol{x}_t, t) dt, \; t \in [0, 1],
\end{equation}
where \( \boldsymbol{v}_\theta \) is the velocity field that guides the flow from \( \boldsymbol{x}_0 \) to \( \boldsymbol{x}_1 \) and \(\boldsymbol{x}_t = t \boldsymbol{x}_1 + (1 - t) \boldsymbol{x}_0 \) represents the linear interpolation between \( \boldsymbol{x}_0 \) and \( \boldsymbol{x}_1 \). To estimate the velocity field \( \boldsymbol{v}_\theta \), CFM minimizes the following loss function:
\begin{equation}
    \centering
    \begin{aligned}
        & \mathcal{L} = \int_0^1 \mathbb{E}_{{\boldsymbol{x}_0},{\boldsymbol{x}_1}}\left[ \left\| \boldsymbol{x}_1 - \boldsymbol{x}_0 - \boldsymbol{v}_\theta(\boldsymbol{x}_t, t) \right\|^2 \right] dt.
    \label{eq:ODE}
    \end{aligned}
\end{equation}
The learned velocity field \( \boldsymbol{v}_\theta \) transports data along simple paths, eliminating the need for intricate time discretization.

\noindent \textbf{State Space Models.}
State Space Models (SSMs) represent a class of sequence models based on continuous systems, which map a one-dimensional input sequence $x(t)$ to $y(t)$ through an implicit latent state $h(t)$. The transformation in an SSM relies on the state matrix $\mathbf{A}$, input matrix $\mathbf{B}$, and output matrix $\mathbf{C}$. The continuous SSM can be described as follows:
\begin{equation}
    h'(t)=\mathbf{A}h(t)+\mathbf{B}x(t); \; y(t)=\mathbf{C}h(t).
\end{equation}
Recent advances in SSMs~\cite{mamba} discretize continuous systems by introducing a timescale parameter $\Delta$, which transforms the continuous parameters $\mathbf{A}$ and $\mathbf{B}$ into their discrete counterparts $\mathbf{\bar{A}}$ and $\mathbf{\bar{B}}$. The discretization is typically performed using the zero-order hold (ZOH), defined by the following expressions:
\begin{equation}
\begin{cases}
\begin{aligned}
    \mathbf{\bar{A}} &= \exp(\Delta \mathbf{A}), \\
    \mathbf{\bar{B}} &= (\Delta \mathbf{A})^{-1} (\exp(\Delta \mathbf{A}) - \mathbf{I}) \cdot \Delta \mathbf{B}.
\end{aligned}
\end{cases}
\end{equation}
The state update process in the discretized system can be formally expressed as:
\begin{equation}
    h_t = \mathbf{\bar{A}} h_{t-1} + \mathbf{\bar{B}} x_t;\; y_t = \mathbf{C} h_t.
\end{equation}
The discretization of SSMs not only preserves the continuous system's ability to model long-range dependencies but also enables efficient computation with linear complexity, making them scalable for processing long sequences.

\subsection{Overview}
\label{4_Overview}
Learning from Demonstration (LfD) provides a structured approach for robots to learn sophisticated skills through expert demonstrations~\cite{schaal1996learning,pomerleau1988alvinn}. This approach employs a dataset \( \mathcal{D}=\{\zeta_i, l_i\}_{i=1}^{N_\mathcal{D}} \) with \({N_\mathcal{D}}\) demonstrations \( \zeta \) coupled with task-specific language instructions \(l\), similar to established methodologies~\cite{wuunleashing,james2022q,shridhar2023perceiver}. The term `\textit{keyframes}' is defined in this context as critical intermediate poses of the end-effector that capture bottleneck actions, identified when joint velocities are near zero and the gripper state remains unchanged, transforming continuous action prediction into discrete keyframe selection~\cite{james2022coarse,johns2021coarse,team2024octo}. Specifically, a demonstration \( \zeta \) is divided into several keyframes, with a total number of \({N_{\zeta}}\), represented by \( \zeta=\{\mathbf{o}_i, \mathbf{a}_i\}_{i=1}^{N_{\zeta}} \), \( \mathbf{o} \) denotes a set of RGB-D images from one or more camera views, while \( \mathbf{a} \) includes the 6-DoF end-effector pose in each keyframe, as shown below:
\begin{equation}
    \mathbf{a} = \{\mathbf{a}_{\text{pos}} \in \mathbb{R}^3, \mathbf{a}_{\text{rot}} \in \mathtt{SO(3)}, \mathbf{a}_{\text{open}} \in \{0, 1\}\},
\end{equation}
where \(\mathbf{a}_{\text{pos}} \) and \(\mathbf{a}_{\text{rot}} \) represent the position and rotation of the end effector, and \(\mathbf{a}_{\text{open}}\) represents the state of the binary gripper. Notably, we utilize a 6D rotation representation to circumvent the discontinuities inherent to quaternion representations~\cite{zhou2019continuity,ke20243d}.


\subsection{Region-Aware Perception}
\label{4_Region-Aware 3D Perception}

\textbf{Dynamic Radius Schedule.}
In this paper, we propose a novel Dynamic Radius Scheduler (DRS), which leverages the flow matching generation process for adaptive 3D perception. The DRS dynamically adjusts the region to focus, allowing the model to flexibly balance between global observation and local geometric detail, ensuring precise perception and manipulation in task-relevant regions. Specifically, DRS outputs a series of region-aware masks \(\mathcal{M}_i\), defined as follows:
\begin{equation}
    \mathcal{M}_i = \left\{ (\boldsymbol{p}_i, r_i) \mid i = 0 \dots N \right\},
\end{equation}
where \( i \) represents the index of the discrete time steps, \( \boldsymbol{p}_i \) represents the noise-perturbed position at time step \(i\), and \(r_i\) is the corresponding region radius. During the inference stage, the noise-perturbed position \( p_i \) gradually converges towards the ground-truth position of the end-effector \( \boldsymbol{p}_0 \). Simultaneously, DRS dynamically modulates the perception radius \(r_i=f(i)\), where \(f(i)\) is a monotonically decreasing function. As the time step \(t\) increases, the radius \(r\) decreases accordingly, following the relationship: \( r_{min} \leq r_i < r_{i-1} < \ldots < r_0\),  which allows the model to change from global perception to focusing on the geometric details of the regions relevant to the task. In our implementation, the radius is controlled by a linear radius control strategy, \( r_i=(1-i) \cdot (r_0 - r_\text{min})  + r_\text{min} \). However, to meet various manipulation requirements, \( f(i) \) can be adjusted to more complex forms, such as cosine annealing, to increase adaptability.

Unlike conventional methods, which typically employ Farthest Point Sampling (FPS) over the global space~\cite{qi2017pointnet}, DRS achieves flexible geometric granularity. DRS not only processes global information, but also preserves finer geometric details in the task-relevant region. During the sampling process, the model can dynamically adjust the resolution based on \( \mathcal{M}_i \). Task-relevant regions are sampled with a higher resolution, as illustrated in \cref{fig:pipeline}, demonstrating how DRS enables FlowRAM to dynamically adjust the granularity of geometric details, significantly improving the capture of geometric features in manipulated regions. 


\noindent \textbf{Multimodal Token Serialization.}
FlowRAM processes both point cloud and RGB images as input, utilizing a dedicated visual encoder for each modality. In particular, for the point cloud branch, we first apply DRS to dynamically sample the required point cloud, and then use a Mamba-based point cloud encoder~\cite{liang2024pointmamba} to generate geometric features \( \boldsymbol{F}_\text{geo} \in \mathbb{R}^{N_1 \times C} \), where \(N_1\) and \(C\) represent the number of sampled points and dimension of the points, respectively. Specifically, \(N_1\) denotes the adaptively sampled points from task-relevant regions, reflecting flexible geometric granularity. For the image branch, we use a CLIP image encoder~\cite{radford2021learning} with feature pyramid network~\cite{lin2017feature} to process multi-view images and extract global semantic features \( \boldsymbol{F}_\text{rgb} \in \mathbb{R}^{ N_2 \times C} \), with \( N_2 \) indicating the number of feature points extracted from the RGB images. In addition, we map the task instruction to \(\boldsymbol{F}_\text{text}\in \mathbb{R}^{ N_3 \times C}\) with a CLIP language encoder, where \( N_3 \) represents the number of tokens extracted from the instructions, and the noise-perturbed pose is projected into \(\boldsymbol{F}_\text{open} \in \mathbb{R}^{1 \times C}\), where \(C\) is the number of feature channels. 

\noindent \textbf{State Space Fusion Model.}
We introduce State Space Models (SSMs) as a multimodal fusion module to enable more effective interaction among noise-perturbed pose, language, semantic and geometric tokens. Before feature fusion, we implement position embedding with a lightweight MLP to encode position information in the unordered point cloud data, thereby enabling the model to efficiently learn 3D corresponding relations. Subsequently, the aforementioned tokens are concatenated along the point dimension for further processing.
\begin{equation}
    \boldsymbol{F}_{\text{in}}=\mathtt{concat}(\boldsymbol{F}_\text{geo}, \boldsymbol{F}_\text{rgb}, \boldsymbol{F}_\text{text}, \boldsymbol{F}_\text{open}),
\end{equation}
where \(\boldsymbol{F}_\text{in} \in \mathbb{R}^{(N_1+N_2+N_3+1) \times C}\). Next, we design a multimodal Mamba model for \(\boldsymbol{F}_\text{in}\) processing, which leverages Mamba's powerful sequence modeling capability to effectively fuse multimodal tokens. Based on a selection mechanism~\cite{mamba}, the model focus on key information and avoid redundant computations, enabling efficient data modeling. The structure of Mamba is shown in~\cref{fig:pipeline}, and its output can be represented by the following equation:
\begin{equation}
    \begin{aligned}
        \boldsymbol{H_1} &= \mathtt{LN}(\boldsymbol{F}_\text{in}) \\
        \boldsymbol{H_2} &= \mathtt{SSM}(\mathtt{Silu}(\mathtt{Conv1d}(\mathtt{Linear}(\boldsymbol{H_1})))) \\
        \boldsymbol{F}_\text{out} &= \mathtt{Linear}(\boldsymbol{H_2} \odot \mathtt{Silu}(\mathtt{Linear}(\boldsymbol{H_1}))),
    \end{aligned}
\end{equation}
where \(\odot\) represents the element-wise product. Equipped with the multimodal Mamba model, FlowRAM efficiently integrates global semantic features with geometric features, preserving geometric details and enhancing precise perception in task-relevant regions.



\subsection{Flow-based Policy Learning}
\label{4_Flow Matching for Robot Policy Learning}

Despite the introduced SSMs enhance efficiency, rapid inference remains challenging due to the iterative nature of diffusion policy. Therefore, we extend CFM for robotic visuomotor policy learning. Let \( \boldsymbol{x}_1 \sim \pi_1\), \( \boldsymbol{x}_0 \sim \pi_0\) on \(\mathbb{R}^{9}\), where \(\pi_1\) represents a multimodal action distribution, while \(\pi_0\) is a standard Gaussian distribution. CFM leverages an ODE to establish a mapping from \(\pi_0\) to \(\pi_1\), creating a path through the interpolation formula
\(\boldsymbol{x}_t = t \boldsymbol{x}_1 + (1 - t) \boldsymbol{x}_0 \).
This equation represents the linear transition from the starting noise distribution toward the target distribution \(\pi_1\) at each time step \(t\), with the target velocity field defined as flows:
\begin{equation}
    u(\boldsymbol{x}_t)=\frac{d\boldsymbol{x}_t}{dt}=\boldsymbol{x}_1-\boldsymbol{x}_0.
\end{equation}
CFM reformulates the mapping as an optimization problem by regressing a parameterized velocity field \(\boldsymbol{v}_{\theta}\) to approximate the target velocity field, facilitating the generation of samples that align with \(\pi_1\). The parameterized velocity prediction network \(v_\theta(\boldsymbol{x}_t,t,\mathcal{C})\) is instantiated as SSMs~\cite{mamba} with MLPs, incorporating adaptive layer normalization~\cite{peebles2023scalable} to inject the conditioning information \( \mathcal{C}\). These conditions include fused features from the state space fusion model, task-specific instructions, and robot proprioception. And the MLP serves as a decoder to decode the velocity field guiding for next keyframe pose generation.

During training, we employed the Logit-Normal Sampling strategy~\cite{esser2024scaling} for time step samlping. CFM optimize \(v_\theta\) to predict the velocity field by minimizing \(\mathcal{L}_\text{CFM}\):
\begin{equation}
    \mathcal{L_\text{CFM}} = \mathbb{E}_{t,\boldsymbol{x}_0, \boldsymbol{x}_1} \left[ \left\| \boldsymbol{x}_1 - \boldsymbol{x}_0 - \boldsymbol{v}_\theta(\boldsymbol{x}_t, t, \mathcal{C}) \right\|^2 \right].
    \vspace{-0.5ex}
\end{equation}
Simultaneously, binary cross-entropy loss supervise the prediction of the end-effector state, encoded by an MLP:
\begin{equation}
    \mathcal{L_{\text{open}}} = -\mathbb{E}_{x,\hat{x}} \left[ x \log(\hat{x}) + (1 - x) \log(1 - \hat{x}) \right],
    \vspace{-0.5ex}
\end{equation}
where \( x \) and \( \hat{x} \) represent the ground-truth state and the predicted state of the gripper, respectively. The final loss is the weighted summation of \(\mathcal{L_\text{CFM}}\) and \(\mathcal{L_\text{open}}\).

\begin{figure*}
    \centering
    \includegraphics[width=1\linewidth]{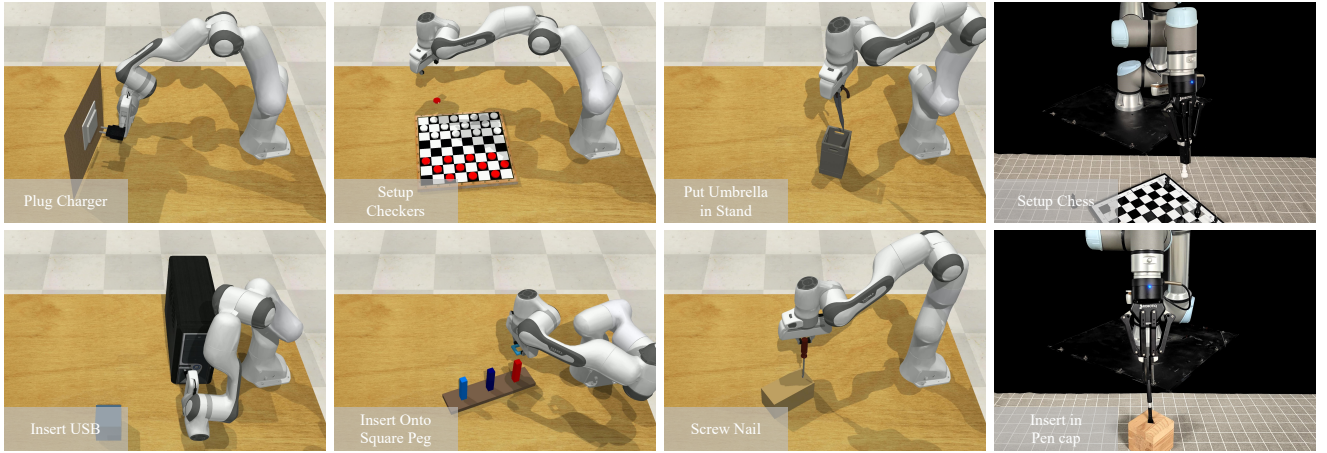}
    \caption{\textbf{Overview of the tasks.} Six high-precision tasks from RLBench and two real-world tasks are visualized. Notably, the \texttt{Unplug Charger} is the inverse process of the \texttt{Plug Charger} in simulator.}
    \label{fig:precise-tasks}
    \vspace{-2ex}
\end{figure*}

In the inference phase, a noisy action pose is first sampled from the source distribution and then transformed into the destination keyframe action by integrating the flow from \(t = 0\) to \(t = 1\) over several steps. Formally, the inference process is defined as follows:
\begin{equation}
    \boldsymbol{x}_{t+\Delta t}=\boldsymbol{x}_t+v_{\theta}(\boldsymbol{x}_t,t,\mathcal{C})\Delta t,
\end{equation}
The \(\boldsymbol{x}_1\) after integration represents the predicted end-effector pose for the next keyframe.


\section{Experiments}
\label{sec:Experiments}

\subsection{Experiment Setting}
\label{sec:experimental setup}
\textbf{Dataset and Setup.}
All simulated experiments are conducted on RLBench~\cite{james2019rlbench}, a challenging large-scale benchmark based on CoppelaSim~\cite{rohmer2013v}. In this environment, a 7-DoF Franka Panda robot with a parallel gripper is employed to perform a range of tasks. We adopt a widely recognized multi-task manipulation setting~\cite{lu2024manigaussian,shridhar2023perceiver,ze2023gnfactor}, which comprises 10 tasks, each with 20 demonstrations and at least two variations, resulting in a total of 166 variations. These variations encompass a multitude of categories, including alterations in shape and color. With limited demonstrations, the agent must develop generalizable manipulation skills to handle varied tasks and environments, rather than overfitting to specific demonstrations. Furthermore, to validate the distinctive capabilities of FlowRAM, we selected 7 high-precision tasks from the RLBench tasksuites. Each task comprises 100 demonstrations captured by 4 RGB-D cameras situated at the front, left shoulder, right shoulder, and wrist of the robot, respectively. The input images have a resolution of 128 \(\times\) 128, and we employ RLBench's native BiRRT motion planner to reach the predicted keyframe pose. A visualization of the task is shown in~\cref{fig:precise-tasks}, while detailed descriptions can be found in the supplementary.

\noindent \textbf{Baselines.}
All methods compared on RLBench utilize 3D information. We contrast FlowRAM with the previous state-of-the-art: PerAct~\cite{shridhar2023perceiver} voxelizes the 3D workspace and employs a perceiver transformer. GNFactor~\cite{ze2023gnfactor} jointly optimizes a generalizable neural field to enhance 3D understanding. ManiGaussian~\cite{lu2024manigaussian} adopts a dynamic Gaussian Splatting framework to model scene-level spatiotemporal dynamics. Act3D~\cite{gervet2023act3d} applies adaptive resolution 3D point sampling to generate hierarchical resolution 3D action maps. 3D Diffuser Actor~\cite{ke20243d} unifies diffusion policies and 3D scene representations, leveraging a 3D denoising transformer to predict action sequences \footnote{All baselines and FlowRAM are trained and tested with input images of size 128 \(\times\) 128, while Act3D and 3D Diffuser Actor use 256 \(\times\) 256.}. RVT-2~\cite{goyal2024rvt2} employs a multi-stage inference pipeline and leverages virtual viewpoint rendering to capture detailed 3D scene information.

\noindent \textbf{Implementation Details.}
We use the AdamW~\cite{loshchilov2017decoupled} optimizer with a learning rate of \( 1e^{-4}\). Additionally, we train FlowRAM for 300K steps using a batch size of 320 and apply EMA on the weights. To fairly compare the performance against previous state-of-the-arts~\cite{goyal2024rvt2,ke20243d}, we reproduce them and train our model with the same GPU type (NVIDIA RTX 3090) and number of GPUs (8). We evaluate the inference speed of diffusion-based methods and FlowRAM by running predictions on the same input data using the same GPU (NVIDIA RTX 3090).
\subsection{Simulation Results}
\label{sec:Simulation Results}

\renewcommand{\arraystretch}{1.1} 
\definecolor{lightgreen}{HTML}{EDF9E6}

\begin{table*}[]

    \centering
    \resizebox{1.0\linewidth}{!}{
        \setlength{\tabcolsep}{0.5em}%
        \begin{tabular}{l|c|cccccccccc}
        \toprule 
        Method / Task & \begin{tabular}[c]{@{}c@{}}Avg.\\ Success↑\end{tabular} & \begin{tabular}[c]{@{}c@{}}Close\\ Jar\end{tabular} & \begin{tabular}[c]{@{}c@{}}Open\\ Drawer\end{tabular} & \begin{tabular}[c]{@{}c@{}}Sweep to\\ Dustpan\end{tabular} & \begin{tabular}[c]{@{}c@{}}Meat off\\ Grill\end{tabular} & \begin{tabular}[c]{@{}c@{}}Turn\\ Tap\end{tabular} & \begin{tabular}[c]{@{}c@{}}Slide\\ Bolck\end{tabular} & \begin{tabular}[c]{@{}c@{}}Put in\\ Drawer\end{tabular} & \begin{tabular}[c]{@{}c@{}}Drag\\ Stick\end{tabular} & \begin{tabular}[c]{@{}c@{}}Put\\ Buttons\end{tabular} & \begin{tabular}[c]{@{}c@{}}Stack\\ Blocks\end{tabular} \\ 
        \midrule 
        PerAct~\cite{shridhar2023perceiver} & 20.4 & 18.7$^{\pm 13.6}$ & 54.7$^{\pm 18.6}$ & 0.0$^{\pm 0.0}$ & 40.0$^{\pm 17.0}$ & 38.7$^{\pm 6.8}$ & 18.7$^{\pm 13.6}$ & 2.7$^{\pm 3.3}$ & 5.3$^{\pm 3.5}$ & 18.7$^{\pm 12.4}$ & 6.7$^{\pm 1.9}$ \\
        GNFactor~\cite{ze2023gnfactor} & 31.7 & 25.3$^{\pm 6.8}$ & 76.0$^{\pm 5.7}$ & 28.0$^{\pm 15.0}$ & 57.3$^{\pm 18.9}$ & 50.7$^{\pm 8.2}$ & 20.0$^{\pm 15.0}$ & 0.0$^{\pm 0.0}$ & 37.3$^{\pm 13.2}$ & 18.7$^{\pm 10.0}$ & 4.0$^{\pm 3.3}$ \\
        
        ManiGaussian~\cite{lu2024manigaussian} & 44.8 & 28.0 & 76.0 & 64.0 & 60.0 & 56.0 & 24.0 & 16.0 & 92.0 & 20.0 & 12.0 \\
        
        Act3D~\cite{ze2023gnfactor} & 65.3 & 52.0$^{\pm 5.7}$ & 84.0$^{\pm 8.6}$ & 80.0$^{\pm 9.8}$ & 66.7$^{\pm 1.9}$ & 64.0$^{\pm 5.7}$ & \textbf{100.0}$^{\pm 0.0}$ & 54.7$^{\pm 3.8}$ & 86.7$^{\pm 1.9}$ & 64.0$^{\pm 1.9}$ & 0.0$^{\pm 0.0}$ \\ 
                
        RVT-2$^{\dagger}$~\cite{goyal2024rvt2} & 76.2 & 79.3$^{\pm 5.3}$ & 78.7$^{\pm 2.7}$ & 87.3$^{\pm 6.7}$ & 86.7$^{\pm 6.7}$ & 85.3$^{\pm 2.9}$ & 76.7$^{\pm 9.5}$ & 86.7$^{\pm 6.7}$ & 96.0$^{\pm 0.0}$ & 67.3$^{\pm 1.1}$ & 27.7$^{\pm 4.0}$ \\
        
        3D Diffuser Actor~\cite{ke20243d} & 78.4 & 82.7$^{\pm 1.9}$ & 89.3$^{\pm 7.5}$ & \textbf{94.7$^{\pm 1.9}$} & \textbf{88.0}$^{\pm 5.7}$ & 80.0$^{\pm 8.6}$ & 92.0$^{\pm 0.0}$ & 77.3$^{\pm 3.8}$ & 98.7$^{\pm 1.9}$ & 69.3$^{\pm 5.0}$ & 12.0$^{\pm 3.7}$ \\

        \midrule 
        \rowcolor{lightgreen} FlowRAM(ours) & \textbf{82.3} & \textbf{85.0}$^{\pm 3.7}$ &\textbf{90.0}$^{\pm 3.3}$ & {88.0}$^{\pm 5.6}$ & 82.0$^{\pm 1.9}$ & \textbf{86.3}$^{\pm 7.7}$ & 93.3$^{\pm 1.9}$ & \textbf{88.0}$^{\pm 5.6}$ & \textbf{100.0}$^{\pm 0.0}$ & \textbf{80.3}$^{\pm 3.7}$ & \textbf{31.3}$^{\pm 2.8}$ \\
        
        \bottomrule  
        \end{tabular}
        }
    \caption{\textbf{Evaluation of Multi-Task on RLBench.} 
    We report the success rate for 25 episodes per task across 10 tasks, along with the average success rate across all tasks. In addition, we provide the mean and standard deviation of the success rates (in \%) on average in 3 random seeds. Variances are included when available. $\dagger$: Model retrained and tested with the same demonstrations as FlowRAM, following the previous setting~\cite{goyal2024rvt2}.}
    \vspace{-2ex}
\label{tab: mani10}
\end{table*}

\definecolor{lightgreen}{HTML}{EDF9E6}
\renewcommand{\arraystretch}{1.1} 

\begin{table}[]
    \centering
    \resizebox{\linewidth}{!}{
        \setlength{\tabcolsep}{0.5em} 
        \begin{tabular}{lllll}
        \toprule 
        \multicolumn{1}{l|}{Method / Task} & \multicolumn{1}{c|}{\begin{tabular}[c]{@{}c@{}}Avg.\\ Success↑\end{tabular}} & \multicolumn{1}{c}{\begin{tabular}[c]{@{}c@{}}Screw\\ Nail\end{tabular}} & \multicolumn{1}{c}{\begin{tabular}[c]{@{}c@{}}Insert onto\\ Square Peg\end{tabular}} & \multicolumn{1}{c}{\begin{tabular}[c]{@{}c@{}}Plug\\ Charger\end{tabular}} \\
        \midrule 
        \multicolumn{1}{l|}{Act3D~\cite{gervet2023act3d}} & \multicolumn{1}{c|}{20.2} & \multicolumn{1}{c}{28.0$^{\pm 3.3}$} & \multicolumn{1}{c}{6.7$^{\pm 3.7}$} & \multicolumn{1}{c}{15.3$^{\pm 2.3}$} \\

        \multicolumn{1}{l|}{RVT-2~\cite{goyal2024rvt2}} & \multicolumn{1}{c|}{39.4} & \multicolumn{1}{c}{50.7$^{\pm 8.5}$} & \multicolumn{1}{c}{53.1$^{\pm 6.0}$} & \multicolumn{1}{c}{34.7$^{\pm 7.1}$} \\

        \multicolumn{1}{l|}{3D Diffuser Actor~\cite{ke20243d}} & \multicolumn{1}{c|}{40.0} & \multicolumn{1}{c}{48.0$^{\pm 1.6}$} & \multicolumn{1}{c}{45.9$^{\pm 2.6}$} & \multicolumn{1}{c}{30.7$^{\pm 5.3}$} \\
        \midrule 
        \rowcolor{lightgreen}
        \multicolumn{1}{l|}{FlowRAM(ours)} & \multicolumn{1}{c|}{\textbf{52.0}} & \multicolumn{1}{c}{\textbf{54.7}$^{\pm 1.9}$} & \multicolumn{1}{c}{\textbf{69.3}$^{\pm 7.5}$} & \multicolumn{1}{c}{\textbf{52.0}$^{\pm 8.6}$} \\ 
        \midrule 

        \midrule 
        \multicolumn{1}{l|}{Method / Task} & \multicolumn{1}{c}{\begin{tabular}[c]{@{}c@{}}Setup \\Checkers \end{tabular}} & \multicolumn{1}{c}{\begin{tabular}[c]{@{}c@{}}Insert\\ USB\end{tabular}} & \multicolumn{1}{c}{\begin{tabular}[c]{@{}c@{}}Unplug\\ Charger\end{tabular}} & \multicolumn{1}{c}{\begin{tabular}[c]{@{}c@{}}Put umbrella\\ in Stand\end{tabular}} \\
        \midrule 
        \multicolumn{1}{l|}{Act3D~\cite{gervet2023act3d}} & \multicolumn{1}{c}{37.3$^{\pm 6.8}$} & \multicolumn{1}{c}{10.3$^{\pm 5.7}$} & \multicolumn{1}{c}{37.3$^{\pm 5.3}$} & \multicolumn{1}{c}{6.7$^{\pm 4.7}$} \\
        \multicolumn{1}{l|}{RVT-2~\cite{goyal2024rvt2}} & \multicolumn{1}{c}{62.7$^{\pm 5.3}$} & \multicolumn{1}{c}{21.3$^{\pm 9.2}$} & \multicolumn{1}{c}{45.3$^{\pm 6.1}$} & \multicolumn{1}{c}{8.0$^{\pm 0.0}$} \\
        \multicolumn{1}{l|}{3D Diffuser Actor~\cite{ke20243d}} & \multicolumn{1}{c}{46.7$^{\pm 3.3}$} & \multicolumn{1}{c}{47.7$^{\pm 1.6}$} & \multicolumn{1}{c}{44.7$^{\pm 5.3}$} & \multicolumn{1}{c}{16.0$^{\pm 3.3}$} \\

        \midrule 
        \rowcolor{lightgreen}
        \multicolumn{1}{l|}{FlowRAM(ours)} & \multicolumn{1}{c}{\textbf{66.7}$^{\pm 6.1}$} & \multicolumn{1}{c}{\textbf{57.3}$^{\pm 3.2}$} & \multicolumn{1}{c}{\textbf{46.7}$^{\pm 1.9}$} & \multicolumn{1}{c}{\textbf{17.3}$^{\pm 9.4}$} \\ 
        \bottomrule  

    \end{tabular}
    }
    \caption{\textbf{Evaluation on High-precision Task on RLBench.} We report the mean and standard deviation of the success rates on average in 3 random seeds. FlowRAM outperforms the previous state-of-the-art method on all tasks.
    }
    \label{tab:precise}
    \vspace{-2ex}
\end{table}

\textbf{Multi-Task Performance.}
As shown in~\cref{tab: mani10}, FlowRAM establishes a new state-of-the-art with an average success rate of 82.3\% across all 10 tasks, achieving the highest success rate in most tasks. Compared to the multi-stage SOTA method~\cite{goyal2024rvt2}, FlowRAM achieves notable improvements in tasks like \texttt{Open Drawer} (+11.3\%) and \texttt{Put Buttons} (+13.0\%). Although the image resolution used in~\cite{ke20243d} is higher than ours, its average success rate remains lower. In particular, for \texttt{Stack Blocks} task, which requires precise pick and place locations, our method significantly outperforms~\cite{ke20243d} by 19.3\%. 

\noindent \textbf{High-precision Task Performance.}
To validate the effectiveness of FlowRAM for high-precision tasks, we analyze it on 7 high-precision tasks~\cite{guhur2023instruction}, comparing it to two multi-stage models, Act3D and RVT-2, as well as a diffusion-based model, 3D Diffuser Actor. The results in~\cref{tab:precise} demonstrate that FlowRAM achieves the highest average success rate of 52.0\% and outperforms all previous methods by a wide margin across all tasks. We find that in \texttt{Insert USB} task, RVT-2 showed limited success, likely due to challenges with severe occlusion. In contrast, FlowRAM, utilizes a dynamic radius schedule, to overcome this occlusion problem by maintaining flexible geometric granularity.

\subsection{Ablation Studies}
\label{sec:Ablation Studies}
\textbf{Analysis of Architecture Performance and Module Contributions.}
Following~\cite{guhur2023instruction,lu2024manigaussian}, we conduct an ablation study in the multi-task setting, with the results presented in~\cref{tab:ablation}. Our main findings are as follows: (1) In terms of architecture, the performance of both is comparable. However, Mamba achieves faster inference speeds than Transformer, owing to its linear computational complexity, especially when geometric information tokens are incorporated via point cloud encoders. (2) The lack of Dynamic Radius Scheduling (DRS), which leads to globally sampled point cloud encoding, not only increases inference time but also reduces model performance. In contrast, when DRS is applied to sample task-relevant regions, strong performance is achieved even when only semantic features are sampled without point cloud encoding. This underscores the importance of task-relevant region perception in the field of robotic manipulation. (3) When the time step is set to 50, changing the training objective to DDIM results in a slight performance decline. In the next section, we will further investigate the impact of the time step on these two training objectives.

\renewcommand{\arraystretch}{1.05} 

\definecolor{darkgreen}{HTML}{25C445}
\definecolor{darkred}{HTML}{DC143C}
\definecolor{lightgrey}{HTML}{E8E8E8}
\definecolor{final}{HTML}{62EF7E}
\definecolor{base}{HTML}{BDBDBD}
\begin{table}[]
    \centering
    \resizebox{1.0\linewidth}{!}{
        \setlength{\tabcolsep}{1.3em} 
        \begin{tabular}{c|ccc|c|c}
        \toprule
        Architecture & DRS & \begin{tabular}[c]{@{}c@{}} P.E. \end{tabular} & \begin{tabular}[c]{@{}c@{}} T.O. \end{tabular} & \begin{tabular}[c]{@{}c@{}}Avg.Success\end{tabular} & \begin{tabular}[c]{@{}c@{}}Inference Time\end{tabular} \\ 
        \midrule
         & {\cellcolor{lightgrey}\color{darkred} \XSolidBrush} & \cellcolor{lightgrey}{\color{darkred} \XSolidBrush} & \cellcolor{lightgrey}CFM & \cellcolor{lightgrey}77.4 & \cellcolor{lightgrey}0.73 \\
         & {\color{darkred} \XSolidBrush} & {\color{darkgreen} \Checkmark} & CFM & 79.1 & 0.85 \\
        Mamba & {\color{darkgreen} \Checkmark} & {\color{darkred} \XSolidBrush} & CFM & 81.5 & 0.75 \\
         & \cellcolor{lightgreen}{\color{darkgreen} \Checkmark} & \cellcolor{lightgreen}{\color{darkgreen} \Checkmark} & \cellcolor{lightgreen}CFM & \cellcolor{lightgreen}\textbf{82.3} & \cellcolor{lightgreen}\textbf{0.87} \\ 
         & {\color{darkgreen} \Checkmark} & {\color{darkgreen} \Checkmark} & DDIM & 82.0 & 0.88 \\ 
         \midrule
         
         & {\color{darkred} \XSolidBrush} & {\color{darkred} \XSolidBrush} & CFM & 77.1 & 0.81 \\ 
         & {\color{darkred} \XSolidBrush} & {\color{darkgreen} \Checkmark} & CFM & 79.3 & 0.94 \\
        Tranformer & {\color{darkgreen} \Checkmark} & {\color{darkred} \XSolidBrush} & CFM & 80.7 & 0.80 \\
         & {\color{darkgreen} \Checkmark} & {\color{darkgreen} \Checkmark} & CFM & 82.1 & 0.97 \\ 
         & {\color{darkgreen} \Checkmark} & {\color{darkgreen} \Checkmark} & DDIM & 81.7 & 0.98 \\ 

         \bottomrule
        \end{tabular}
    }
    \caption{\textbf{Comparison of Architectures and Modules.} We present a comparison between the Mamba and Transformer architectures, tested under various configurations combining Dynamic Radius Scheduling (DRS), point encoder~\cite{liang2024pointmamba}, and different training objectives (CFM vs. DDIM). All models are evaluated with \(i\) = 50 inference steps, with inference time reported in seconds (s). Note: baseline (in \textcolor{base}{\(\blacksquare\)}) and final (in \textcolor{final}{\(\blacksquare\)}).}
    \label{tab:ablation}
    \vspace{-3ex}
\end{table}

\begin{figure*}
    \centering
    \includegraphics[width=1\linewidth]{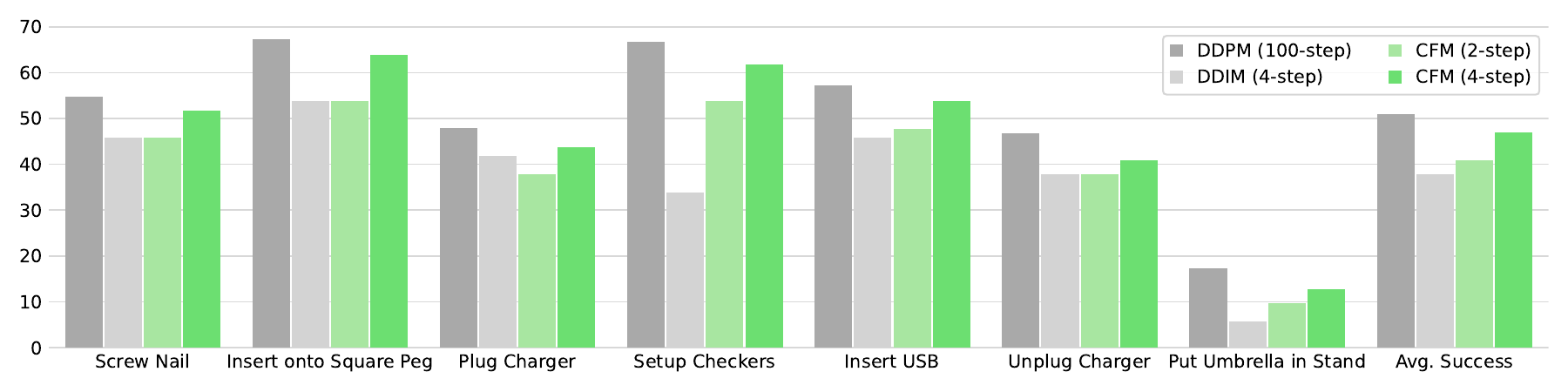}
    \caption{\textbf{Performance of CFM in High-Precision Tasks.} We compare FlowRAM with DDPM and DDIM across 7 high-precision tasks, results demonstrate that CFM achieves strong performance in just 2 steps, showing excellent results even in high-precision scenarios.}
    \label{fig:ab_task_CFMDDPM}
    \vspace{-2ex}
\end{figure*}

\noindent \textbf{Analysis of the Impact of Time Steps.}
To clearly analyze the impact of the inference step \(i\) on the performance and inference speed, and to explore whether CFM is a more effective training objective compared to more established diffusion models, we compare DDPM~\cite{ho2020denoising} with 100 inference steps, DDIM~\cite{song2020denoising} and CFM with different inference steps \(i\). As shown in~\cref{tab:flowvsddim}, CFM outperforms DDIM with fewer time steps. Notably, with only 2 steps, DDIM fails to generate feasible actions for task completion, while CFM maintains a high success rate, making it more suitable for training objective. Additionally, to illustrate the advantages of CFM in high-precision tasks, ~\cref{fig:ab_task_CFMDDPM} presents the success rate across various tasks, demonstrating the applicability of CFM in a wide range of scenarios.

\definecolor{lightgreen}{HTML}{EDF9E6}

\renewcommand{\arraystretch}{1.05} 

\begin{table}[]
    \centering
    \resizebox{1.0\linewidth}{!}{
        \setlength{\tabcolsep}{0.7em}
        \begin{tabular}{lc|ccccc}
        \toprule 
         & \(i\)-steps & 2 & 4 & 8 & 16 & 32 \\ 
        \midrule 
        \begin{tabular}[c]{@{}l@{}}DDIM\\CFM\end{tabular} & \begin{tabular}[c]{@{}c@{}} Avg.\\Success(↑) \end{tabular} & \begin{tabular}[c]{@{}c@{}}29.6\\\cellcolor{lightgreen}\textbf{74.2}\end{tabular} & \begin{tabular}[c]{@{}c@{}}69.8\\\cellcolor{lightgreen}\textbf{77.8}\end{tabular} & \begin{tabular}[c]{@{}c@{}}77.6\\\cellcolor{lightgreen}\textbf{78.6}\end{tabular} & \begin{tabular}[c]{@{}c@{}}\cellcolor{lightgreen}\textbf{79.8}\\79.3\end{tabular} & \begin{tabular}[c]{@{}c@{}}80.9\\\cellcolor{lightgreen}\textbf{81.1}\end{tabular} \\ 
        \midrule 
        \begin{tabular}[c]{@{}l@{}}DDIM\\CFM\end{tabular} & \multicolumn{1}{l|}{\begin{tabular}[c]{@{}l@{}} Inference\\Time (↓) \end{tabular}} & \begin{tabular}[c]{@{}c@{}}58.7\\\cellcolor{lightgreen}\textbf{57.7} \end{tabular} & \begin{tabular}[c]{@{}c@{}}92.0\\\cellcolor{lightgreen}\textbf{91.0}\end{tabular} & \begin{tabular}[c]{@{}c@{}}140.3\\\cellcolor{lightgreen}\textbf{139.7}\end{tabular} & \begin{tabular}[c]{@{}c@{}}249.3\\\cellcolor{lightgreen}\textbf{248.5}\end{tabular} & \begin{tabular}[c]{@{}c@{}}\cellcolor{lightgreen}\textbf{495.7}\\496.3\end{tabular} \\ 
        \bottomrule  
        \end{tabular}
    }

    \caption{\textbf{Impact of inference steps.} We conducted tests using the same tasks, measuring the average inference time (in \textbf{ms}) across all episodes. The results show that CFM and DDIM exhibit similar inference times. However, CFM demonstrates superior performance when the number of inference steps is reduced.
    }
    \label{tab:flowvsddim}
\end{table}

\renewcommand{\arraystretch}{1.0} 


\definecolor{lightgreen}{HTML}{EDF9E6}

\begin{table}[t]
    \centering
    \resizebox{1.0\linewidth}{!}{
    \setlength{\tabcolsep}{0.2em}
        \begin{tabular}{l|cc|c}
        \toprule
        Task & \multicolumn{1}{c}{\# Variation} & \multicolumn{1}{c|}{\# Train traj.} & \multicolumn{1}{c}{Success(↑)} \\ \midrule
        Put fruit in bowl       & object (4)  & 12 & \cellcolor{lightgreen}\textbf{10/10} \\
        Place Earphone on block & object (2)  & 10 & \cellcolor{lightgreen}\textbf{10/10} \\
        Take Skewer to tiny box & color (2) & 6 & \cellcolor{lightgreen}\textbf{8/10} \\
        Setup Chess             & color and object (4) & 12 & \cellcolor{lightgreen}\textbf{8/10}\\
        Sort Shape              & N/A (1)  & 6  & \cellcolor{lightgreen}\textbf{7/10} \\
        Insert in Pen cap          & N/A (1)   & 10 & \cellcolor{lightgreen}\textbf{6/10} \\ \midrule
        All tasks               & 14      & 56 & \cellcolor{lightgreen}\textbf{49/60} \\ \bottomrule
        \end{tabular}
    }
    \caption{\textbf{Real-world tasks configuration and performance.} We conducte experiments on 6 real-world tasks, evaluating 10 episodes for each task and reporting the success rate.
    }
    \label{tab:real-robot}
    \vspace{-2ex}
\end{table}

\begin{figure}
    \centering
    \includegraphics[width=1.0\linewidth]{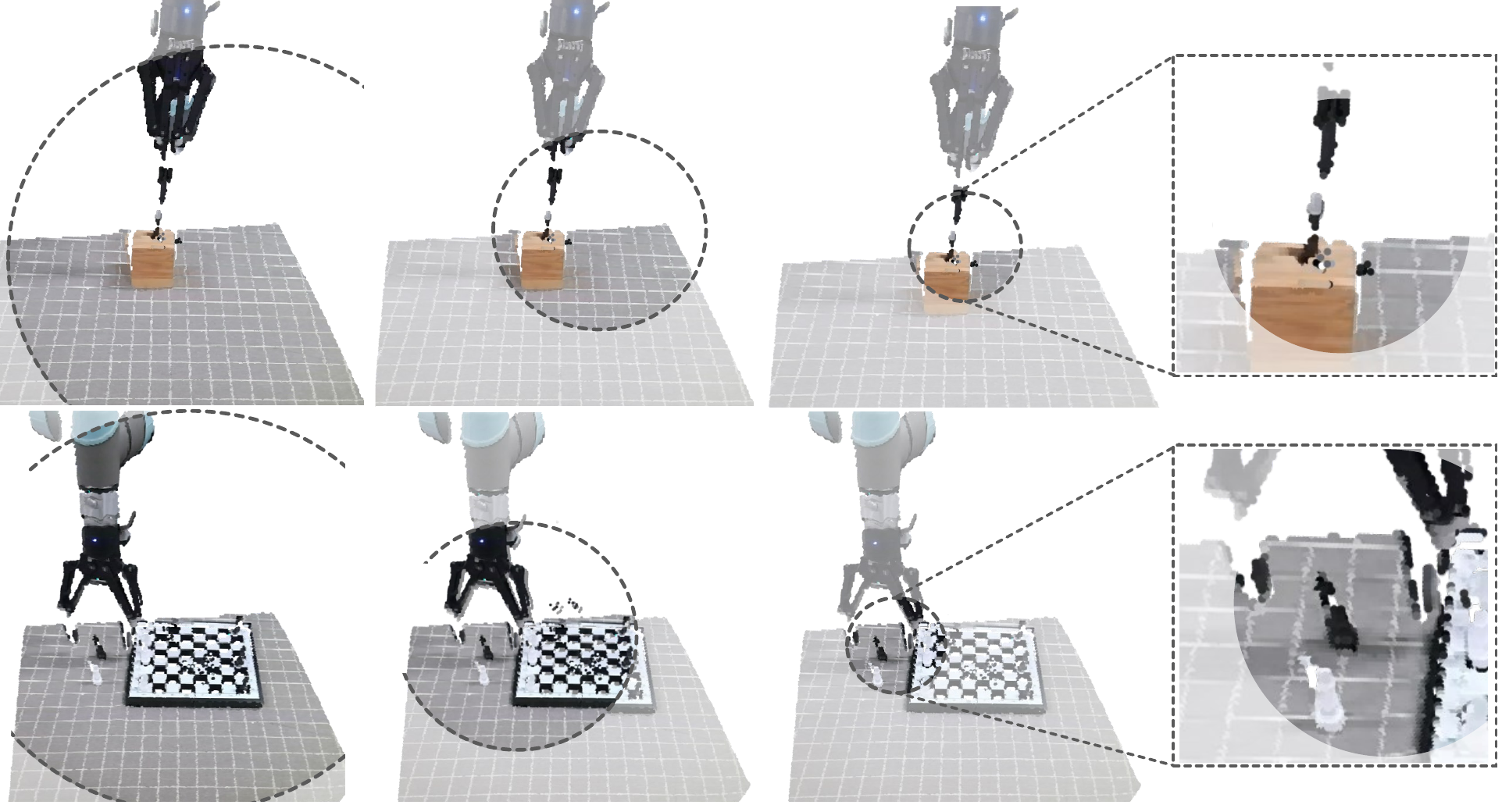}
    \caption{\textbf{Perception Region Illustration.} We visualize different regions controlled by the dynamic radius schedule in real-world.}
    \label{fig:real_world}
    \vspace{-2ex}
\end{figure}
\subsection{Real-Robot Results}
We conduct a series of real robot experiments using a 6-DoF UR5 robotic arm equipped with a Robotiq 2F-140 two-finger gripper and a single Azure Kinect (RGB-D) camera in a front view. Our experimental setup includes 6 language-conditioned tasks with 14 variations, including different objects and colors. We collected 56 demonstration trajectories by executing human-defined waypoints. Images are initially captured at a resolution of 2048 × 1536 and downsampled to 256 × 192. For inference, we apply the BiRRT planner integrated with MoveIt! ROS package~\cite{coleman2014reducing} to reach the predicted action poses. Visual examples of our tasks can be found in~\cref{fig:precise-tasks}, with quantitative results presented in~\cref{tab:real-robot}. Meanwhile, we visualize the results of the dynamic radius schedule in the real-world task in~\cref{fig:real_world}. These results show that FlowRAM effectively learns precise, real-world manipulation tasks from a limited set of demonstrations, and achieves reliable task execution.



\section{Conclusion}
\label{sec:Conclusion}

In this paper, we present FlowRAM, a novel framework that combines region-aware 3D perception with conditional flow matching for robotic manipulation. We design a dynamic radius schedule exploiting the flow matching generation process to seamlessly integrate global information and fine geometric details, while efficiently handling multimodal feature fusion with state space models. Extensive experiments demonstrate competitive performance and remarkable inference speed of FlowRAM on various robotic tasks. We envision our work will inspire future research on flow-based policies and feature fusion in robotics. 







\section*{Acknowledgement}

This work was supported in part by the National Key Research and Development Project under Grant 2024YFB4708100, National Natural Science Foundation of China under Grants 62088102, U24A20325 and 12326608, and Key Research and Development Plan of Shaanxi Province under Grant 2024PT-ZCK-80.

{
    \small
    \bibliographystyle{ieeenat_fullname}
    \bibliography{main}
}

\clearpage
\setcounter{page}{1}
\maketitlesupplementary


In this supplementary material, we provide additional details and experiments not included in the main paper due to space limitations.

\renewcommand{\thefigure}{A}
\begin{figure*}[t]
    \centering
    \includegraphics[width=1\linewidth]{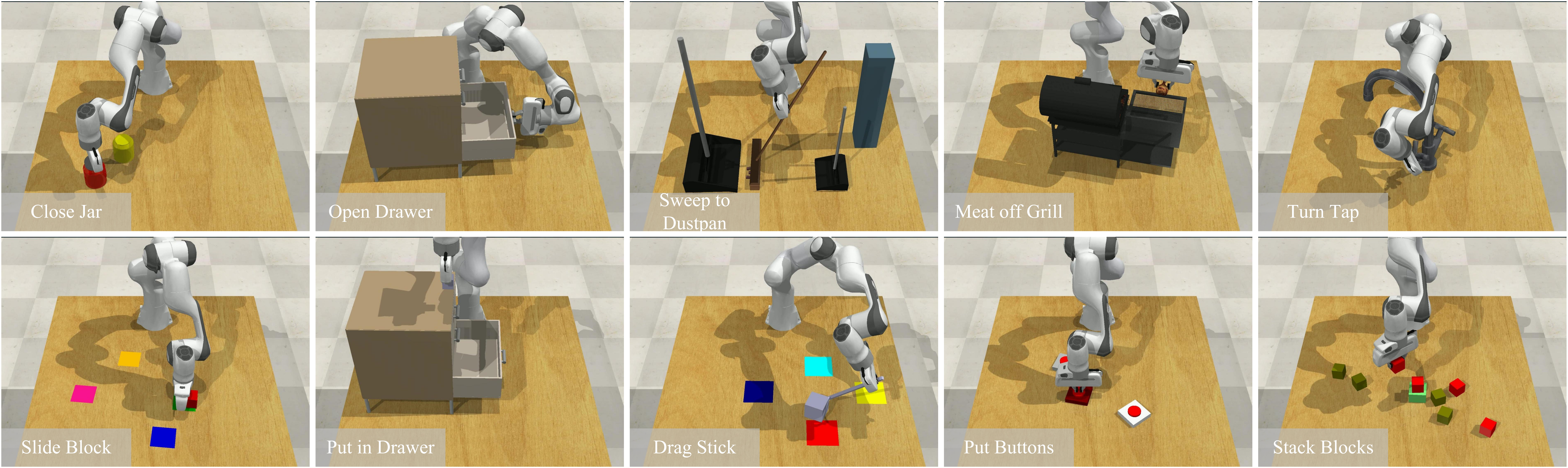}
    \caption{\textbf{Additional Task Visualization.}We visualize all tasks in the single-view setting.}
    \label{sup:10task}
    \vspace{-2ex}
\end{figure*}

\section*{A. RLBench Tasks and Training Pipeline}
\label{sec:Tasks and Pipeline}
\subsection*{A.1 Details of RLBench Tasks}
We present two experimental settings employed in our work. The first involves 10 tasks from RLBench~\cite{james2019rlbench}, summarized in~\cref{sup:task_describe} and visualized in~\cref{sup:10task}. Each task encompasses at least two or more variations to evaluate the multi-task capabilities of the agent. Notably, due to the templated nature of the instructions, which vary with each variation, the agent learns a language-guided multi-task policy, rather than learning one-off policies for single variation tasks~\cite{shridhar2023perceiver}. Additionally, we evaluate FlowRAM on high-precision tasks by selecting representative tasks from the RLBench benchmark based on~\cite{guhur2023instruction}.

\renewcommand{\thetable}{A}
\renewcommand{\arraystretch}{1.08} 
\definecolor{lightgreen}{HTML}{EDF9E6}

\begin{table}[H]
    \centering
    \resizebox{1.0\linewidth}{!}{
        \setlength{\tabcolsep}{0.2em} 
        \begin{tabular}{l|ccl}
        \toprule
        Task                                  & \# Variation     & \# Keyframe & Instruction Template                                         \\ \midrule
        Close Jar                             & color(20)        & 6.0          & `close the \_ jar'                           \\
        Open Drawer                           & placement(3)     & 3.0          & `open the \_ drawer'                       \\
        Sweep to Dustpan & size(2)          & 4.6          & `sweep dirt to dustpan'                              \\
        
        Meat off Grill                        & object(2)        & 5.0          & `take the \_ off the grill'                     \\
        Turn Tap                              & placement(2)     & 2.0          & `turn \_ tap'                                \\
        Slide Block                           & color(4)         & 4.7          & `slide the block to \_ target'                    \\
        Put in Drawer                         & placement(3)     & 12.0         & `put the item in the \_ drawer'                  \\
        Drag Stick                            & color(20)        & 6.0          & `use stick to drag cube onto the \_ target' \\
        Put Buttons                           & color(50)        & 3.8          & `push the \_ button'                                \\
        Stack Blocks                          & color, count(60) & 14.6         & `stack to \_\_ blocks'                            \\
        \bottomrule
        \end{tabular}
    }
    \caption{\textbf{single-view setting Tasks.} We introduce variations of these tasks, the number of keyframes and instructions template. }
    
    \label{sup:task_describe}
    \vspace{-2ex}
\end{table}

\subsection*{A.2 Details of Training Pipeline}
To facilitate policy learning, we uniformly sample a set of expert episodes across all task variations. From these sampled episodes, we randomly select input-action pairs for each task to construct a training batch. Similar to previous work~\cite{gervet2023act3d,shridhar2023perceiver,goyal2024rvt2,guhur2023instruction}, the agent is assumed to use a sampling-based motion planner, which helps define input-action pairs as keyframes, as described in~\cref{4_Overview}. \textit{This setup simplifies the original sequential decision-making problem into predicting the next optimal keyframe action based on the current observation. }

\section*{B. Details of FlowRAM}

\definecolor{lightgreen}{HTML}{EDF9E6}

\renewcommand{\thetable}{B}

\begin{table}[]
    \centering
    \resizebox{0.8\linewidth}{!}{
    \setlength{\tabcolsep}{2.0em}
    \begin{tabular}{c|c}
        \toprule
        \multicolumn{1}{l|}{\textbf{Hyperparameter}} & \textbf{Value}         \\ \midrule
        \multicolumn{1}{l|}{\textbf{Training}}       &               \\ 
        batch size                         & 320           \\
        training iteration                 & 300K          \\
        learning rate                      & \( 1e^{-4}\)      \\
        weight decay                       & \( 5e^{-4}\)      \\
        optimizer                          & AdamW         \\
        EMA                                & 0.9999  \\
        gradient moment            &  (0.9, 0.999)           \\
        loss weight : \(\mathcal{L}_\text{CFM}\)            & 100           \\
        loss weight : \(\mathcal{L}_\text{open}\)               & 10            \\ \midrule
        
        \multicolumn{1}{l|}{\textbf{Model}}          &               \\
        image resolution                   & 128 \(\times\) 128        \\
        embedding dim  \(C\)                    & 120           \\
        noise scheduler                    & EulerDiscrete \\
        FlowMatching timestep              & 50            \\
        sampled semantic tokens \(N_1\)      & 4096          \\
        sampled geometric tokens \(N_2\)   & 1024          \\ \midrule
        
        \multicolumn{1}{l|}{\textbf{Mamba}}          &               \\
        d\_model                           & 120           \\
        d\_state                           & 16            \\
        d\_conv                            & 4             \\ 
        expand                             & 2             \\ \bottomrule
        
        \end{tabular}
    }
    \caption{\textbf{Hyper-parameters of FlowRAM.}}
    \label{sup:parameter}
    \vspace{-3ex}
\end{table}

The hyperparameters used in FlowRAM are shown in \cref{sup:parameter}. Specifically, \(\mathbf{a}_{\text{pos}} \in \mathbb{R}^3\), we use a 6D rotation representation to avoid the inherent discontinuities of quaternions, where \(\mathbf{a}_{\text{rot}} \in \mathtt{R}^6\). We apply a Flow Matching Discrete Scheduler for adding noise to both the position \( \mathbf{a}_{\text{pos}}\) and the rotation \( \mathbf{a}_{\text{rot}}\).
To provide a clear understanding of the flow matching policy, we provide pseudocode for flow-based model training~\cref{sup:1} and inference~\cref{sup:2}.

\section*{C. Additional Experiment and Visualization}
\subsection*{C.1 Additional Experiment}
To further evaluate the proposed Dynamic Radius Schedule (DRS), we analyze its effect over a few time steps, with qualitative results presented in~\cref{sup:ablation_DRS}.

\definecolor{lightgreen}{HTML}{EDF9E6}
\renewcommand{\thetable}{C}

\begin{table}[H]
    \centering
    \resizebox{1.0\linewidth}{!}{
    \setlength{\tabcolsep}{0.8em}
        \begin{tabular}{l|ccccc|c}
        \toprule
        \(i\)-steps & 2    & 4    & 8     & 16    & 32    &                \\
        \midrule
        CFM &   70.5   &   73.6   & 74.0  &   74.6  &  76.5   & Avg. Sucess     \\
        w/o DRS  &    60.7  &   94.3   &    143.1   &  252.0   & 499.7      & Inference Time \\
        \midrule
        CFM    & \cellcolor{lightgreen}\textbf{74.2} & \cellcolor{lightgreen}\textbf{77.8} &\cellcolor{lightgreen}\textbf{78.6}  & \cellcolor{lightgreen}\textbf{79.3}  & \cellcolor{lightgreen}\textbf{81.1}  & Avg. Sucess     \\
        w DRS        & \cellcolor{lightgreen}\textbf{57.7} & \cellcolor{lightgreen}\textbf{91.0}   & \cellcolor{lightgreen}\textbf{139.7} & \cellcolor{lightgreen}\textbf{248.5} & \cellcolor{lightgreen}\textbf{496.3} & Inference Time \\ 
        \bottomrule
        \end{tabular}
    }
    \caption{\textbf{Ablation study about the DRS in few timesteps.} Note: If DRS is not used, it means that the global point cloud is downsampled to extract geometric features.
    }
    \label{sup:ablation_DRS}
    \vspace{-2ex}
\end{table}

The incorporation of DRS consistently enhances the success rate of the model across all time steps. For example, at \(i=2\), the success rate improves from 70.5\% (w/o DRS) to 74.2\% (w/ DRS), and at \(i=32\), from 76.5\% to 81.1\%, confirming that DRS contributes to higher accuracy and robustness, especially at larger inference steps, while also reducing inference time. The results demonstrate that in robotic manipulation tasks, leveraging local information significantly enhances task success rates. The DRS mechanism dynamically adjusts the radius, enabling the model to focus on task-relevant regions efficiently while minimizing redundant global computations.

\begin{algorithm}
\caption{Conditional Flow Matching Policy Training in Euclidean Space}
\label{sup:1}
\begin{algorithmic}[1]
\REQUIRE Dataset \( \zeta=\{\boldsymbol{o}, \mathtt{a}\} \), Instructions \(l\), Observations $\boldsymbol{o}$, Conditions $\mathcal{C}=\{\boldsymbol{o}, l\}$, Actions $\mathtt{a}= \{\mathtt{a}_{\text{pos}}, \mathtt{a}_{\text{rot}}\}$
\REPEAT
    \STATE $\mathtt{a_1}, \boldsymbol{o} \sim \zeta $ \textcolor{gray!70}{\# \textit{Sample a random input-action pair \\ from the dataset}}
    \STATE $\mathtt{a_0} \sim \mathcal{N}(0, 1)$ \textcolor{gray!70}{\# \textit{Sample from a Gaussian distribution}}
    \STATE $t \sim \text{Uniform}\{0, \dots, 1\}$ \textcolor{gray!70}{\# \textit{Sample a random time t}}
    \STATE $\mathtt{a_t} \gets t \cdot \mathtt{a_1} + (1 - t) \cdot \mathtt{a_0}$  \textcolor{gray!70}{\# \textit{Linear interpolation}}
    \STATE $u(\mathtt{a_t}) \gets \mathtt{a_1} - \mathtt{a_0}$  \textcolor{gray!70}{\# \textit{Compute target velocity Field}}
    \STATE $\hat{u}(\mathtt{a_t}) \gets \boldsymbol{v}_\theta(\mathtt{a_t}, t, \mathcal{C})$ \textcolor{gray!70}{\# \textit{Predicted velocity Field}}
    \STATE $\mathcal{L}_{\text{CFM}} \gets \| u(\mathtt{a_t}) - \hat{u}(\mathtt{\mathtt{a_t}}) \|^2$ \textcolor{gray!70}{\# \textit{Compute loss}}
    \STATE $\theta \gets \theta - \eta\nabla_\theta \mathcal{L}_{\text{CFM}}$ 
\UNTIL{Converged}
\end{algorithmic}
\end{algorithm}

\begin{algorithm}
\caption{Action Poses Generation based on Conditional Flow Matching Policy}
\label{sup:2}
\begin{algorithmic}[1]
\REQUIRE Observation $\boldsymbol{o}$, Number of steps \(k_{\text{steps}}\), \\ condition $\mathcal{C} = \{\boldsymbol{o}, l\}$
\STATE \(\mathtt{a}_0 \sim \mathcal{N}(0, 1)\) \textcolor{gray!70}{\# Sample from a Gaussian distribution}
\FOR{$k = 1$ to $k_{\text{steps}}$}
    \STATE \(t \gets k / k_{\text{steps}}\) \textcolor{gray!70}{\# Compute time step}
    \STATE \(\Delta t \gets 1 / k_{\text{steps}}\) \textcolor{gray!70}{\# Compute interval}
    \STATE \(\boldsymbol{P}\gets \boldsymbol{v}_\theta(z, t, \mathcal{C})\) \textcolor{gray!70}{\# Predict velocity}
    \STATE \(\mathtt{a}_t \gets \mathtt{a}_0 + \boldsymbol{P} \cdot \Delta t\) \textcolor{gray!70}{\# Update state}
\ENDFOR
\RETURN $\mathtt{a}_t$
\end{algorithmic}
\end{algorithm}

\renewcommand{\thetable}{D}

\begin{table*}[t]
    \centering

    \resizebox{1.0\linewidth}{!}{
    \setlength{\tabcolsep}{0.5em}%
    \begin{tabular}{l|c|c|c|c|c|c|c|c|c|c}
        \toprule 
        \multicolumn{1}{l|}{Method} & \multicolumn{1}{c|}{\begin{tabular}[c]{@{}c@{}}Avg. \\ Success↑\end{tabular}} & \multicolumn{1}{c|}{\begin{tabular}[c]{@{}c@{}}Push\\ Buttons\end{tabular}} & \multicolumn{1}{c|}{\begin{tabular}[c]{@{}c@{}}Slide\\ Block\end{tabular}} & \multicolumn{1}{c|}{\begin{tabular}[c]{@{}c@{}}Sweep to\\ Dustpan\end{tabular}} & \multicolumn{1}{c|}{\begin{tabular}[c]{@{}c@{}}Meat off\\ Grill\end{tabular}} & \multicolumn{1}{c|}{\begin{tabular}[c]{@{}c@{}}Turn\\ Tap\end{tabular}} & \multicolumn{1}{c|}{\begin{tabular}[c]{@{}c@{}}Put in\\ Drawer\end{tabular}} & \multicolumn{1}{c|}{\begin{tabular}[c]{@{}c@{}}Close\\ Jar\end{tabular}} & \multicolumn{1}{c|}{\begin{tabular}[c]{@{}c@{}}Drag\\ Stick\end{tabular}} & \begin{tabular}[c]{@{}c@{}}Put in\\ Safe\end{tabular} \\
        \midrule 
        PerAct & 49.4 & 92.8 \(^{\pm 3.0}\) & 74.0 \(^{\pm 13.0}\) & 52.0 \(^{\pm 0.0}\) & 70.4 \(^{\pm 2.0}\) & 88.0 \(^{\pm 4.4}\) & 51.2 \(^{\pm 4.7}\) & 55.2 \(^{\pm 4.7}\) & 89.6 \(^{\pm 4.1}\) & 86.0 \(^{\pm 3.2}\) \\
        3D Diffuser Actor & 81.3 & 98.4 \(^{\pm 2.0}\) & 97.6 \(^{\pm 3.2}\) & 84.0 \(^{\pm 4.4}\) & 96.8 \(^{\pm 1.6}\) & 99.2 \(^{\pm 1.6}\) & 96.0 \(^{\pm 3.6}\) & 96.0 \(^{\pm 2.5}\) & 100.0 \(^{\pm 0.0}\) & \textbf{97.6 \(^{\pm 2.0}\)} \\
        RVT-2 & 81.4 & 100.0 \(^{\pm 0.0}\) & 92.0 \(^{\pm 2.8}\) & \textbf{100.0 \(^{\pm 0.0}\)} & \textbf{99.0 \(^{\pm 1.7}\)} & 99.0 \(^{\pm 1.7}\) & \textbf{96.0 \(^{\pm 0.0}\)} & \textbf{100.0 \(^{\pm 0.0}\)} & 99.0 \(^{\pm 1.7}\) & 96.0 \(^{\pm 2.8}\) \\
        \rowcolor{lightgreen} FlowRAM (ours) & \textbf{84.9} & \textbf{100.0 \(^{\pm 0.0}\)} & \textbf{100.0 \(^{\pm 0.0}\)} & 92.0 \(^{\pm 2.0}\) & 94.0 \(^{\pm 2.0}\) & \textbf{100.0 \(^{\pm 0.0}\)} & 92.0 \(^{\pm 0.0}\) & 96.0 \(^{\pm 2.0}\) & \textbf{100.0 \(^{\pm 0.0}\)} & 96.0 \(^{\pm 0.0}\) \\
        \bottomrule
        \toprule
        Method & \begin{tabular}[c]{@{}c@{}}Place\\ Wine\end{tabular} & \begin{tabular}[c]{@{}c@{}}Screw\\ Bulb\end{tabular} & \begin{tabular}[c]{@{}c@{}}Open\\ Drawer\end{tabular} & \begin{tabular}[c]{@{}c@{}}Stack\\ Blocks\end{tabular} & \begin{tabular}[c]{@{}c@{}}Stack\\ Cups\end{tabular} & \begin{tabular}[c]{@{}c@{}}Put in\\ Cupboard\end{tabular} & \begin{tabular}[c]{@{}c@{}}Insert\\ Peg\end{tabular} & \begin{tabular}[c]{@{}c@{}}Sort\\ Shape\end{tabular} & \begin{tabular}[c]{@{}c@{}}Place\\ Cups\end{tabular} &  \\
        \midrule
        PerAct & 44.8 \(^{\pm 7.8}\) & 17.6 \(^{\pm 2.0}\) & 88.0 \(^{\pm 5.7}\) & 26.4 \(^{\pm 3.9}\) & 2.4 \(^{\pm 2.2}\) & 28.0 \(^{\pm 4.4}\) & 5.6 \(^{\pm 4.1}\) & 16.8 \(^{\pm 4.7}\) & 2.4 \(^{\pm 3.2}\) &  \\
        3D Diffuser Actor & 93.6 \(^{\pm 4.8}\) & 82.4 \(^{\pm 2.0}\) & 89.6 \(^{\pm 4.1}\) & 68.3 \(^{\pm 3.3}\) & 47.2 \(^{\pm 8.5}\) & 85.6 \(^{\pm 4.1}\) & 65.6 \(^{\pm 4.1}\) & 44.0 \(^{\pm 4.4}\) & 24.0 \(^{\pm 7.6}\) &  \\
        RVT-2 & 95.0 \(^{\pm 3.3}\) & \textbf{88.0 \(^{\pm 4.9}\)} & 74.0 \(^{\pm 11.8}\) & \textbf{80.0 \(^{\pm 2.8}\)} & \textbf{69.0 \(^{\pm 5.9}\)} & 66.0 \(^{\pm 4.5}\) & 40.0 \(^{\pm 0.0}\) & 35.0 \(^{\pm 7.1}\) & 38.0 \(^{\pm 4.5}\) &  \\
        \rowcolor{lightgreen} FlowRAM (ours) & \textbf{96.0\(^{\pm 0.0}\)} & 84.0\(^{\pm 2.3}\)  & \textbf{92.0 \(^{\pm 0.0}\)} & 77.3 \(^{\pm 3.8}\) & 61.0 \(^{\pm 2.0}\) & \textbf{86.0\(^{\pm 4.0}\)} &\textbf{ 72.0\(^{\pm 2.7}\)} & \textbf{48.0\(^{\pm 4.0}\)} & \textbf{42.0 \(^{\pm 2.3}\)} & \\
        \bottomrule

    \end{tabular}
    }
    \label{tab:18 task on rlbench}
     \caption{\textbf{Evaluation on RLBench with 100 demonstrations.} We report the success rate for 18 RLBench tasks and the average success rate across all the tasks. Additionaly, we show the mean and standard deviation of success rates (in \%) average across three random seeds. FlowRAM outperforms all prior arts on most tasks, especially for tasks that require high geometric understanding.}
    \vspace{-3ex}

\end{table*}

\subsection*{C.2 Flow Matching Generation process}

\cref{sup:Flow Matching Examples} showcases the Flow Matching Generation process of the trained FlowRAM agent in predicting the next keyframe pose, where time step \(i\) is set to 50 for a clear depiction of the Flow Matching Generation process. FlowRAM learns accurate velocity fields across various tasks, which facilitates a more efficient flow matching generation process during inference. (For better visualization, we eliminate some of the noise inherent in the depth camera.)

\renewcommand{\thefigure}{B}
\begin{figure}[H]
    \centering
    \includegraphics[width=1.0\linewidth]{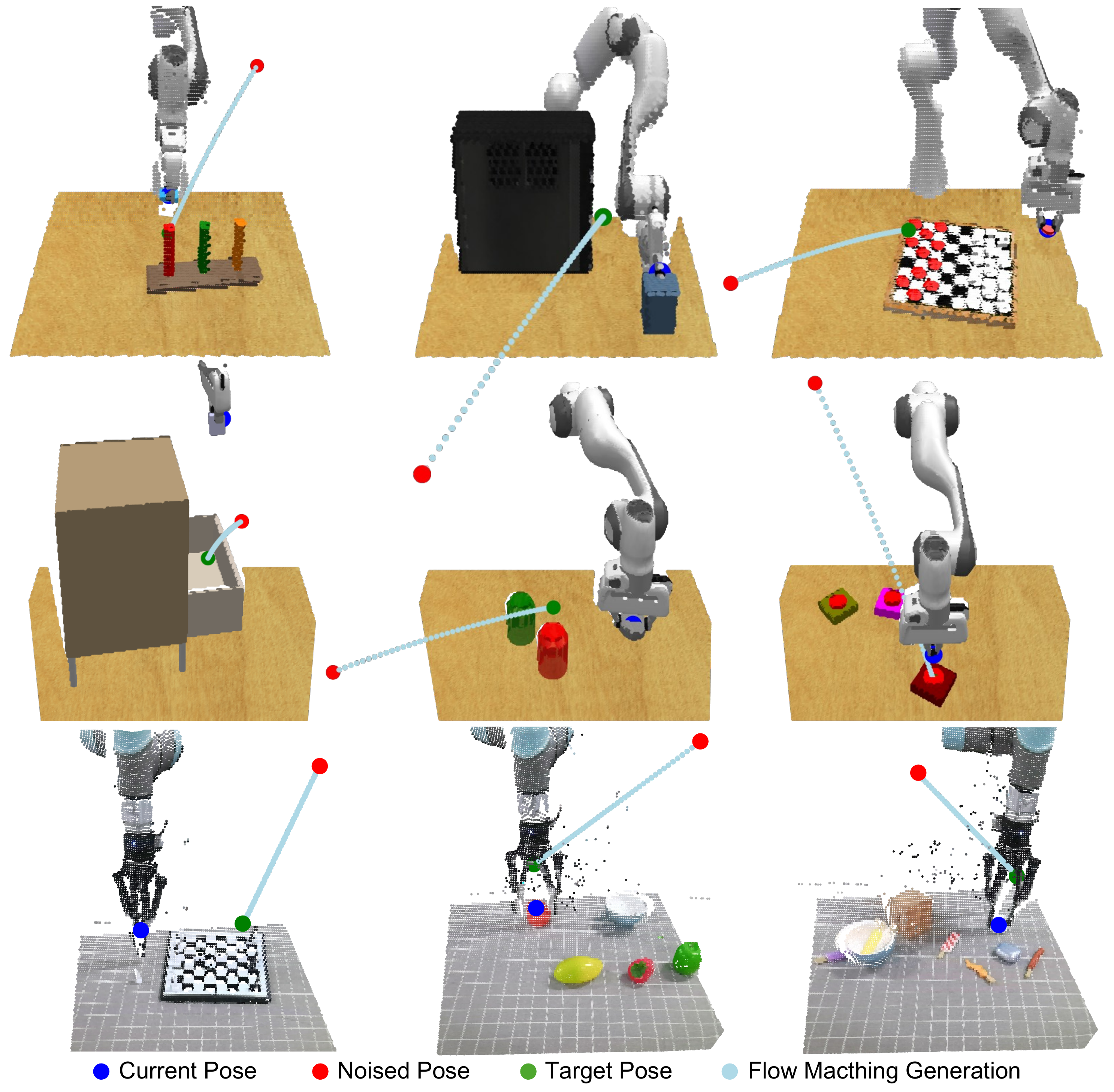}
    \caption{\textbf{Examples of the next keyframe pose prediction using FlowRAM.} The first two rows represent high-precise and single-view simulation tasks, while the third row corresponds to real-world tasks.}
    \label{sup:Flow Matching Examples}
\end{figure}

\subsection*{C.3 Multi-Modal Actions}
Policies based on generative models allow the modeling of multimodal action distributions~\cite{chi2023diffusion}, i.e., scenarios where there are multiple valid actions given observations and instruction. \cref{sup:Multi-Modal Actions} shows a typical example of FlowRAM in multimodal action prediction. With multiple \texttt{red blocks} to choose from, there is diversity in the poses of the next keyframe. Therefore, during training, starting from the noise distribution, FlowRAM effectively learns the multimodal distribution of keyframe actions by fitting the vector field along probabilistic flow paths. For inference, one path is chosen for each noise sample and numerical integration is used to generate the end-effector pose for the next keyframe.

\renewcommand{\thefigure}{C}
\begin{figure}[H]
    \centering
    \includegraphics[width=0.5\linewidth]{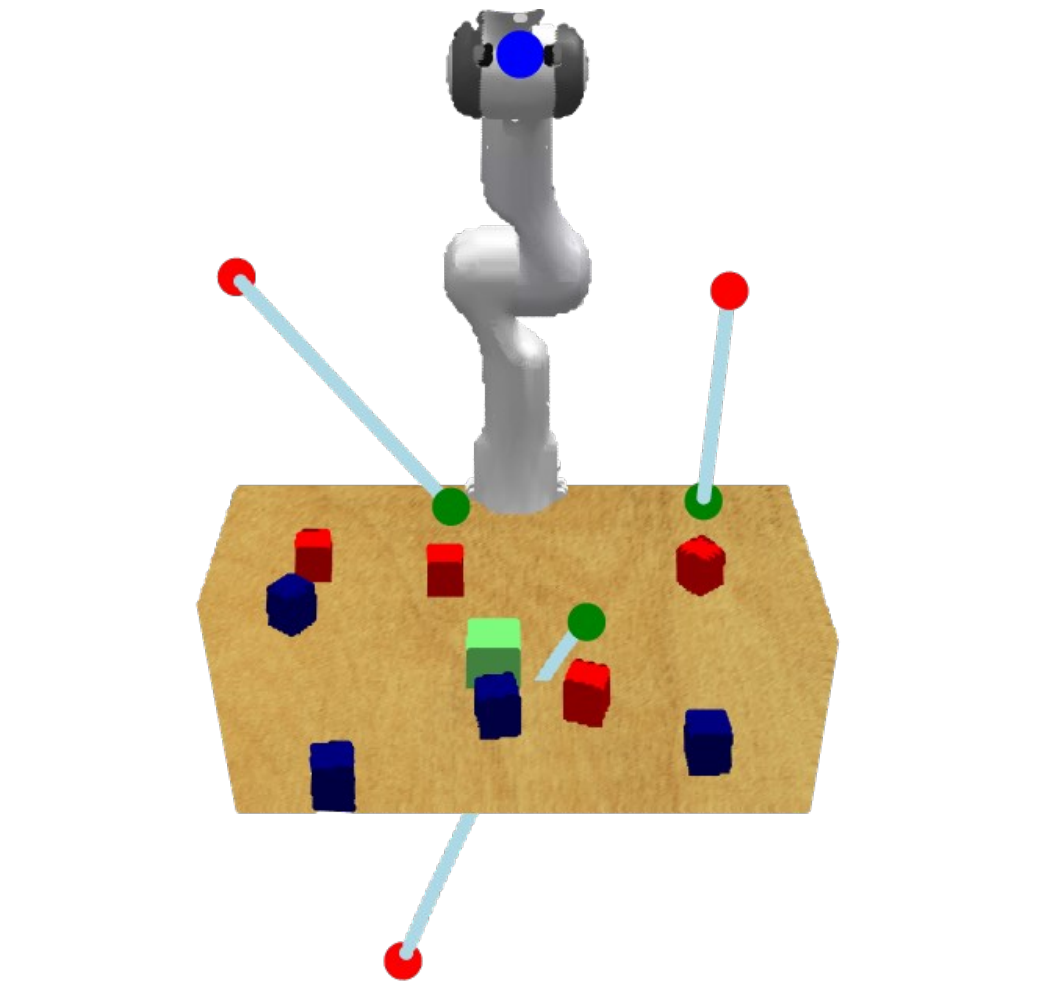}
    \caption{\textbf{Multi-Modal Predictions.}}
    \label{sup:Multi-Modal Actions}
\end{figure}

\section*{D. Experiments on a major benchmark}
We understand the importance of the 18-task benchmark in RLBench for evaluating the generalization and robustness of robotic learning algorithms. Therefore, we provide a comprehensive analysis of the results, including success rates, statistical significance, and insights into the performance of FlowRAM. The success rates for each task, along with the average success rate across all tasks, are summarized in \cref{tab:18 task on rlbench}. FlowRAM demonstrated excellent performance, consistently achieving high success rates across the majority of tasks.

\section*{E. Failure cases and Limitations}
\subsection*{E.1 Analysis of Failure cases}
In the 17 simulation tasks, \texttt{Put umbrella in Stand} task showed the poorest performance. Upon careful examination, we find that while FlowRAM is able to grasp the umbrella handle reliably, two primary errors lead to failure during the insertion process: (1) The umbrella is not correctly inserted into the stand, occurring in approximately 25\% of cases; (2) Although the umbrella is successfully inserted into the stand and has detached from the gripper after the two-finger gripper opens, due to a simulator bug, the umbrella continues to follow the center point of the gripper and is pulled out of the stand, occurring in approximately 55\% of cases.

In \cref{tab: mani10}, FlowRAM fails to achieve state-of-the-art performance on the \texttt{Sweep to Dustpan} and \texttt{Meat off Grill} tasks. Further analysis reveals that at an earlier checkpoint (e.g., 150K training iterations), these tasks achieved success rates of 96\% and 92\%, respectively, while other tasks remained under-optimized. We hypothesize that the reported results in \cref{tab: mani10} may reflect performance trade-offs caused by a uniform task sampling strategy, where equal sampling weights led to performance gains in some tasks at the expense of others~\cite{goyal2024rvt2,shridhar2023perceiver}.

\subsection*{E.2 Limitations}
Although FlowRAM has demonstrated remarkable accuracy and scalability in both simulated and real-world scenarios, it still faces challenges in balancing multi-task learning. Future research could focus on exploring advanced strategies for effectively optimizing multi-task learning while maintaining high accuracy.


\end{document}